\documentclass[preprint,12pt, a4paper]{elsarticle}

\let\today\relax
\makeatletter
\def\ps@pprintTitle{%
	\let\@oddhead\@empty
	\let\@evenhead\@empty
	\def\@oddfoot{\footnotesize\itshape
		{  } \hfill\today}%
	\let\@evenfoot\@oddfoot
}
\makeatother

\usepackage{amssymb}
\usepackage[hyphens]{url}

\def\tool{diff-SAT} 
\def\githubrep{\url{https://github.com/MatthiasNickles/diff-SAT}}

\begin{document}

\begin{frontmatter}

\title{\tool{} - A Software for Sampling and Probabilistic Reasoning for SAT and Answer Set Programming}

\author{Matthias Nickles}

\address{School of Computer Science\\National University of Ireland, Galway\\matthias.nickles@nuigalway.ie}

\begin{abstract}

This paper describes \tool{}, an Answer Set and SAT solver which combines regular solving with the capability to use probabilistic clauses, facts and rules, and to sample an optimal world-view (multiset of satisfying Boolean variable assignments or answer sets) subject to user-provided probabilistic constraints. The sampling process minimizes a user-defined differentiable objective function using a gradient descent based optimization method called Differentiable Satisfiability Solving ($\partial\mathrm{SAT}$) respectively Differentiable Answer Set Programming ($\partial\mathrm{ASP}$). Use cases are i.a. probabilistic logic programming (in form of Probabilistic Answer Set Programming), Probabilistic Boolean Satisfiability solving (PSAT), and distribution-aware sampling of model multisets (answer sets or Boolean interpretations).

\end{abstract}

\begin{keyword}

Boolean Satisfiability \sep Answer Set Programming \sep Probabilistic Satisfiability \sep Probabilistic Programming \sep Gradient Descent 

\end{keyword}

\end{frontmatter}


\section{Introduction}

Modern solvers for the Boolean Satisfiability Problem (SAT) are powerful, versatile and fast tools for logical inference. SAT solvers underlie a large number of other optimization and reasoning approaches as well as industry applications, for example many constraint programming and Satisfiability Modulo Theories (SMT) solvers, bounded model checking, package managers, hardware and software verification tools, crypto-analysis, and planning and scheduling approaches \cite{AshishSabharwal2011,Lafitte2014,MarquesSilva2008}. 
Answer Set Programming (ASP) \cite{stable-model-semantics} is a form of logic programming (with syntax similar to Prolog and Datalog) which is closely related to SAT solving and oriented primarily towards NP-hard search and combinatorial problems. \\

Regular SAT and Answer Set solvers do not have the capability to handle or reason about probabilistic knowledge, or to sample multisets of models governed by specific probability distributions. \tool{} (formerly named delSAT) is an open source SAT and ASP solver which has those capabilities. \tool{} is programmed in Scala and runs on any contemporary standard Java VM. Its solver core incorporates modern SAT solving techniques such as CDCL (Conflict-Driven Clause Learning)-style solving, Glucose-style restart policy \cite{Audemard2009}, adaptive branching heuristics, rephasing, and parallel portfolio solving in multi-core environments. \\

Symbolic knowledge can be provided in various formats, including Boolean formulas in CNF and logic programs in Answer Set Programming (ASP) intermediate format (obtained from an ASP grounder). Soft constraints can be provided by attaching probabilities to clauses, facts or rules, or by specifying differentiable objective functions directly as mathematical expressions. In contrast to most existing logic programming approaches, probabilistic databases and approaches based on a reduction to Weighed Model Counting \cite{CHAVIRA2008772}, \tool{} does not require any independence assumptions about probabilistic variables. Compared to Markov Logic Networks \cite{mlnOlder} (the arguably most popular probabilistic logic), \tool{} allows for probabilistic logic programming and has thus the ability to deal with inductive definitions and to use probabilities directly as weights despite being able to model complex interdependencies.\\
\tool{}'s output is a multiset of models (if the given problem has a solution) corresponding to a probability distribution over possible worlds, allowing for probabilistic inference in the usual way.\\

On the technical level, \tool{} achieves its functionality by finding optimal distributions of entire multi-sets of satisfying models (related to the concept of ``worldviews'' in epistemic logic programming \cite{kahl_et_al:OASIcs:2018:9867}). An incremental sampling process minimizes a user-defined objective function (loss function) over symbol statistics in model multisets using a gradient descent-based branching literal selection approach called Differentiable Satisfiability Solving respectively Differentiable Answer Set Programming \cite{DBLP:conf/ilp/Nickles18aa,Nickles18} ($\partial\mathrm{SAT},\ \partial\mathrm{ASP}$). \\

Probabilistic annotations of logical items like clauses or rules can be automatically translated into such differentiable functions. Multi-models optimization could be stylized as a form of model optimization where the optimal model is a meta-model which comprises multiple models, but \tool{} avoids this stylization as it would require clunky techniques like model reification. Remark: Multi-models optimization should not be confused with other optimization types involving multiple models, such as the concurrent discovery of several optimal models. \\

In contrast to existing approaches to ASP- or SAT-based optimization such as MaxWalkSAT for Weighted MaxSAT solving \cite{Kautz96ageneral}, the differentiable objective function evaluates and optimizes statistics over an entire multiset of models instead of finding an optimal single model which is optimal wrt. the sum of the weights of satisfied soft constraints (soft clauses). Compared to other SAT sampling approaches, in particular \cite{chakraborty-aaai14}, \tool{} samples multiple models using a form of gradient descent until a given minimization threshold of the objective function has been reached. Clauses can be attached probabilities directly instead of non-normalized weights. Note that \tool{} is not a Stochastic Local Search (SLS) solver but a complete solver (i.e., a solver which returns UNSAT if the input formula is not satisfiable) with probabilistic optimization features.\\
A more detailed description of \tool{}'s algorithm and comparison with further related approaches can be found in \cite{Nickles18,DBLP:conf/ilp/Nickles18aa}.

\section{Functionality and Input}

Notwithstanding being based on a single, relatively simple technical principle ($\partial\mathrm{SAT}\ /\ \partial\mathrm{ASP}$), \tool{} goes beyond regular SAT/ASP solving by providing a wide range of additional use cases (besides plain SAT and ASP solving), including probabilistic SAT (PSAT), Probabilistic Answer Set Programming (PrASP), and distribution-dependent model sampling where a family of target distributions can be specified by differentiable objective functions of the form described in the previous section. \\

\noindent \tool{} is open source. The source code, recent releases and documentation can be found at
\githubrep.\\

\noindent Features include:

\begin{itemize}[-]
	\item Parallel portfolio ASP and SAT solving based on Conflict-Driven Nogood Learning (CDNL), a variant of CDCL) 
	\item Runs on the JVM (Java 1.8 or higher)
	\item Model sampling as randomized, gradient-steered search for optimal multisets of models (answer sets respectively Boolean assignments)
	\item Objective functions are arbitrary differentiable functions over atom frequencies in models, allowing for, e.g., probabilistic atoms (clauses, rules) and probabilistic inference, i.e., computation of the probabilities of logical query formulas.	
	\item A range of alternative input formats, including for Probabilistic SAT and Probabilistic Answer Set Programming. See below for the complete list of supported formats.
	\item User API for working with probabilistic and non-probabilistic clause and rule sets 
	\item Direct support for basic non-ground ASP rules (unrestricted non-ground answer set programs can also be used, but that requires a preceding grounding step using a grounder such as Gringo or Clingo \cite{DBLP:series/synthesis/2012Gebser}, as usual with Answer Set solvers)
\end{itemize}

Input is accepted in various standard as well as non-standard forms (where no commonly accepted format exists yet, in particular in the area of probabilistic input):

\begin{itemize}[-]
\item DIMACS CNF (for regular SAT solving)

\item Probabilistic DIMACS CNF (PCNF) where each clause can optionally be annotated with a probability to turn it into a soft clause

\item Enhanced DIMACS CNF with a list of parameter atoms and objective function terms appended to the DIMACS part

\item ASP Intermediate Format (ASPIF), optionally extended with parameter atoms and objective function terms defined using special predicates

\item Enhanced ASPIF format with a list of parameter atoms and objective function terms appended to the ASPIF part

\item Probabilistic ASP Intermediate Format (PASPIF) which enhances ASPIF standard format with optional rule probabilities 
\end{itemize}

\tool{}'s output for satisfiable problems is a multiset of models (possible worlds) where the normalized count of a model represents its probability. The probabilities of individual clauses, facts or rules can be obtained using a simple subsequent filtering and counting step as in most other probabilistic logic programming approaches, but \tool{} can optionally perform this step itself for user-specified query clauses, facts and rules.

\newpage 

\section*{Code metadata} 

\begin{table}[!h]
	\begin{tabular}{|l|p{6.5cm}|p{6.5cm}|}
		\hline
		\textbf{Nr.} & \textbf{Code metadata description} &  \\
		\hline
		C1 & Current code version & 0.5.2 \\
		\hline
		C2 & Link to code/repository used for this code version & \githubrep \\
		\hline
		C4 & Code License   & MIT \\
		\hline
		C5 & Code versioning system used & git \\
		\hline
		C6 & Software code languages, tools, and services used & Scala, Java, JVM \\
		\hline
		C7 & Compilation requirements, operating environments \& dependencies & JDK 1.8+ (e.g., OpenJDK), SBT or Maven\\
		\hline
		C8 & If available link to developer documentation/manual & \githubrep \\
		\hline
		C9 & Support email for questions & (see author contact details)\\
		\hline
	\end{tabular}
	\caption{Code metadata}
	\label{} 
\end{table}


\bibliographystyle{plain} 
\bibliography{NicklesArxiv2020}

\end{document}